\pdfoutput=1

\documentclass[11pt]{article}

\usepackage[]{ACL2023}

\usepackage{times}
\usepackage{latexsym}

\usepackage[T1]{fontenc}

\usepackage[utf8]{inputenc}

\usepackage{microtype}

\usepackage{inconsolata}

\usepackage{graphicx}
\usepackage{booktabs}
\usepackage{pgfplots}
\pgfplotsset{compat=1.10}
\newcommand{\aquamuse}{AQuAMuSe}

\usepackage{amsmath}
\usepackage{booktabs}

\usepackage{wrapfig}
\usepackage{url}
\usepackage{comment}
\usepackage{rotating}
\usepackage{multirow}
\usepackage{float}
\usepackage{colortbl}
\usepackage{pifont} 
\newcommand{\cmark}{\ding{51}} %
\usepackage{arydshln}

\title{Learning to Plan and Generate Text with Citations}

\author{Constanza Fierro\thanks{~~Equal contribution.} \ \thanks{~~Work done during internship at Google.} \ $^{\natural}$ \ \ Reinald Kim Amplayo\footnotemark[1] \ $^{\ddagger}$ \ \ \\
        \textbf{Fantine Huot$^{\ddagger}$ \ \ Nicola De Cao$^{\ddagger}$ \ \ Joshua Maynez$^{\ddagger}$ \ \ Shashi Narayan$^{\ddagger}$ \ \ Mirella Lapata$^{\ddagger}$} \\
        $^{\natural}$Department of Computer Science, University of Copenhagen \\ 
        $^{\ddagger}$Google DeepMind \\ 
        \texttt{c.fierro@di.ku.dk} \\ \texttt{\{reinald,fantinehuot,ndecao,joshuahm,shashinarayan,lapata\}@google.com}}

\begin{document}
\maketitle
\begin{abstract}
  The increasing demand for the deployment of LLMs in
  information-seeking scenarios has spurred efforts in creating
  verifiable systems, which generate responses to queries along with
  supporting evidence. In this paper, we explore the
  \emph{attribution} capabilities of plan-based models which have been
  recently shown to improve the faithfulness, grounding, and
  controllability of generated text. We conceptualize plans as a
  sequence of questions which serve as \emph{blueprints} of the
  generated content and its organization. We propose two attribution
  models that utilize different variants of blueprints, an
  \textit{abstractive} model where questions are generated from
  scratch, and an \textit{extractive} model where questions are copied
  from the input. Experiments on long-form question-answering show
  that planning consistently improves attribution quality. Moreover,
  the citations generated by blueprint models are more accurate
  compared to those obtained from LLM-based pipelines lacking a
  planning component.
\end{abstract}

\section{Introduction}

Large language models (LLMs) have demonstrated remarkable abilities to
engage in creative
conversations~\citep{Thoppilan:ea:2022,openai2023gpt4}, summarize
information from contextual cues
\citep{goyal2022news,zhang2023benchmarking}, and deliver zero-shot
performance on a wide range of previously unseen predictive and
generative tasks \citep{NEURIPS2020_1457c0d6, chowdhery2022palm}. They
are also becoming increasingly useful in information-seeking
scenarios, ranging from answering simple questions
~\citep{roberts-etal-2020-much, rae2021scaling} to generating
responses to search-like queries
\citep{Nakano:ea:2021,menick2022teaching}.

The increasing demand for the deployment of LLMs in
information-seeking scenarios has further spurred efforts in creating
verifiable systems, which generate responses to queries along with
supporting evidence. The evidence can take the form of a URL pointing
to a short segment of text which supports an answer
\citep{bohnet2022attributed}, an attribution report with evidence
snippets \citep{gao2022attributed}, quotes cited verbatim from
pages retrieved from a search engine \citep{menick2022teaching}, and
references to passages extracted while browsing
\citep{Nakano:ea:2021,gao2023enabling}. In fact, this last type of evidence has been
recently adopted in the form of in-line citations by commercial search
engines such as BingChat\footnote{\url{https://bing.com/new}} and
perplexity.ai\footnote{\url{https://perplexity.ai}}.

Regardless of how the evidence is presented, recent approaches tend to
rely on a retrieval system (e.g., a commercial search engine) to
obtain passages relevant to a query, while an LLM conditions on them to
generate a response
\citep{menick2022teaching,Nakano:ea:2021,bohnet2022attributed}.  Other
work generates an answer to the input query first and subsequently
retrieves relevant evidence in a post-processing step
\citep{bohnet2022attributed}. Alternatively, the retrieved evidence can
be used to further revise the generated response rendering it more
consistent with the evidence \citep{gao2022attributed}.

\begin{figure*}[t]
  \includegraphics[width=\textwidth]{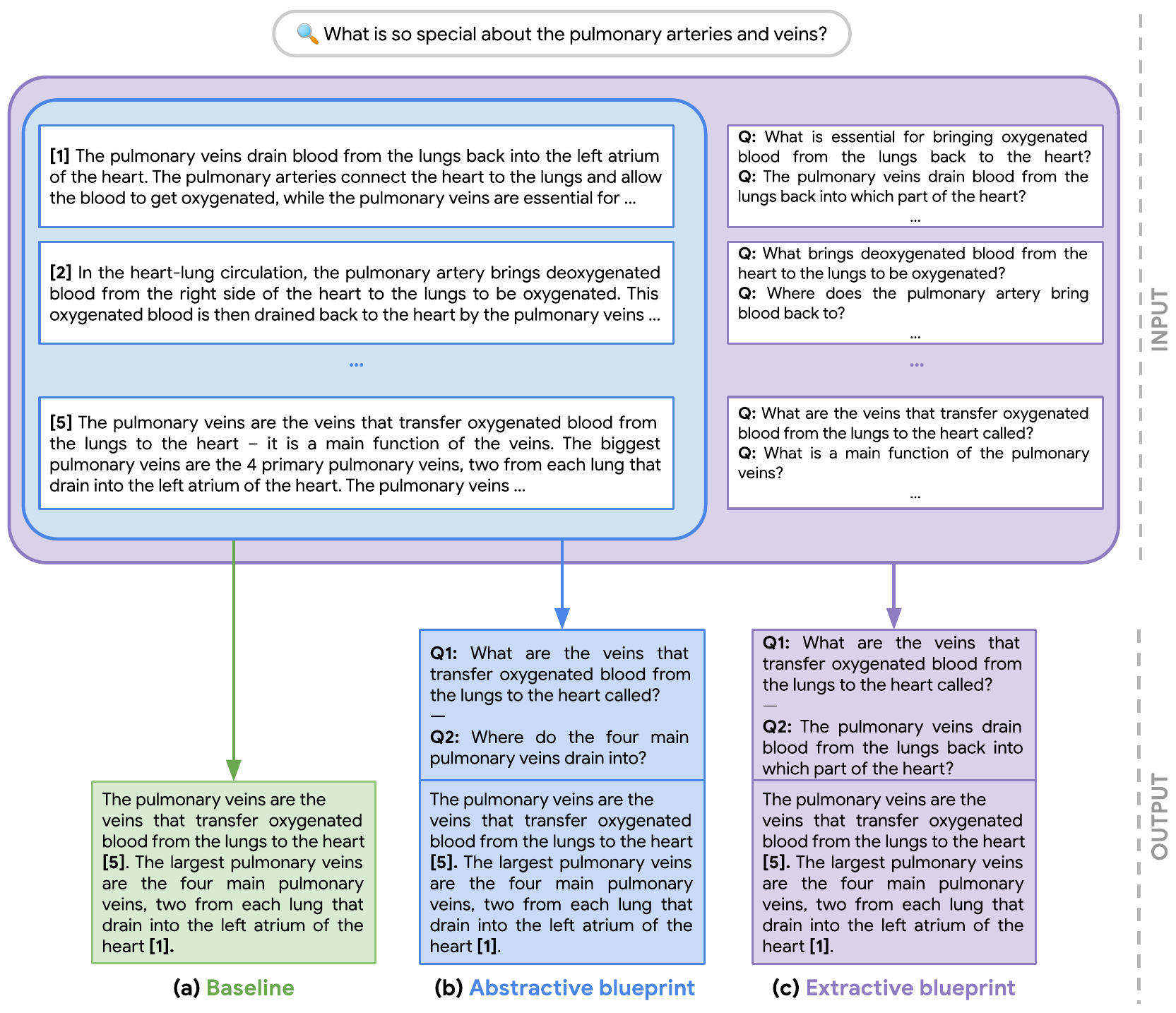}
  \centering
  \vspace*{-.7cm}
  \caption{\label{figure:example} Query (top), followed by most
    relevant (abridged) passages, and summaries (bottom) with in-line
    citations. Summary~(a) is the output of a vanilla
    sequence-to-sequence model trained to generate long answers with
    citations. Summaries~(b) and~(c) are the output of models with
    abstractive and extractive plans, respectively. Citations for
    plan-based models can have different formats (e.g., references to
    the question plan; see Section~\ref{sec:results:analysis}).}
\end{figure*}

Despite recent efforts, it remains an open question how to best
develop models with a \emph{built-in} mechanism for attribution to
external evidence. A related question is whether said mechanism
contributes to generating more factually faithful output. Large-scale
evaluation studies paint a worrying picture. \citet{liu2023evaluating}
find that long-form responses from existing search engines frequently
contain unsupported statements or inaccurate citations, while
\citet{bohnet2022attributed} show that model performance on
attribution varies greatly (between 46\% and 71\%) across different
architectures for the simpler question answering task.

In this paper, we explore the attribution capabilities of plan-based
models which have been shown to be less prone to hallucinations and
more controllable \citep{moryossef-etal-2019-step, puduppully2019data,
  narayan-etal-2021-planning,
  narayan2022conditional,huot-etal-2023-text,huot2023muplan}.  We focus on long-form
question answering
\citep{fan-etal-2019-eli5,vig-etal-2022-exploring,xu-lapata-2022-document}
which aims to generate summaries from a set of passages that answer a
specific query. We simulate how a search engine might synthesize
passages of high relevance to a user query by assuming access to a
retriever, and some way of verifying the output, i.e.,~by citing
sources (see Figure~\ref{figure:example}).  Our models operate on
retrieved passages and learn to plan and generate summaries with
attribution.  On account of being more expressive, plan-based models
allow us to formalize different forms of attribution, e.g.,~plans can
be verified via citations to passages, while summaries can be verified
through citations to the plan, passages, or both.

Our models conceptualize text plans as a sequence of questions
operating as \emph{blueprints} for generation, determining what to say
and in which order
\cite{narayan2022conditional,liu-etal-2023-visual-storytelling,huot-etal-2023-text}. Questions
as a planning mechanism are ideally suited to attribution, since they
provide a natural link between retrieved passages and their
summaries. We define two models which differ on whether the
question-based plan is \emph{generated} (see
Figure~\ref{figure:example}(b)) or \emph{copied} from input passages
(Figure~\ref{figure:example}(c)) and explore whether explicit planning
has any bearing on citation quality. Our contributions can be
summarized as follows:
\begin{itemize}
\itemsep0em 
\item We develop automatic methods to annotate training data with
  plans \emph{and} citations, and fine-tune several Transformer models
  \citep{vaswani:ea:2017} to generate attributed text.\footnote{Our
    data and code are available from \url{xxx.yyy.zzz}.}
\item Experimental results on the AQuAMuSe dataset
  \citep{kulkarni2020aquamuse} demonstrate that plans consistently
  improve attribution quality. Furthermore, 
 summary quality improves with an extractive blueprint model.
\item Out-of-domain experiments on the ALCE benchmark
  \citep{gao2023enabling} show that, once acquired, attribution is a
  robust skill (across information-seeking tasks). In terms of
  attribution quality, our models are competitive with (and sometimes
  better than) pipelines that heavily rely on large language models.

\end{itemize}

%

\section{Related Work}

LLMs have made significant advances in recent
years in generating high quality natural language
responses~\citep{NEURIPS2020_1457c0d6, chowdhery2022palm}. Despite
impressive capabilities, they also demonstrate high levels of opacity:
it is unclear where the information they generate comes from, which can
 undermine their trustworthiness~\citep{cheng-etal-2022-mapping}.

A variety of techniques aim to remedy this by building language models
that provide references to support generated text. The majority of
existing work has focused on models which \emph{learn} to generate
citations. For example, \citet{Nakano:ea:2021} and
\citet{menick2022teaching} use reinforcement learning from human
preferences to train language models to answer questions and provide
supporting evidence (in the form of retrieved
snippets). \citet{bohnet2022attributed}  generate both an
answer and a URL to a web page from which a paragraph is subsequently
selected as support for the answer. They also experiment with
attribution as a post-processing step where no learning is involved
and use retrieval to select sources supporting their model's output.

Along similar lines, \citet{gao2022attributed} propose a two-stage
approach where generated text is first post-edited to be made
attributable (to web content) and then revised accordingly.  In
addition to modeling, efforts have been also directed to the creation
of evaluation protocols and benchmarks for assessing whether a
statement is supported by provided evidence
\citep{rashkin2021measuring,bohnet2022attributed,liu2023evaluating}.

Our work focuses on plan-based models which have been shown to be less
opaque and more controllable \citep{moryossef-etal-2019-step,
  puduppully2019data, narayan-etal-2021-planning,
  narayan2022conditional,huot-etal-2023-text}.  We draw inspiration
from \citet{narayan2022conditional} who formulate plans as a sequence
of question-answer pairs. 
We develop various plan-based models which can perform attribution, and quantify the extent to which the two mechanisms interact and whether they are mutually beneficial. 


\section{Problem Formulation}
\label{sec:formulation}

We follow a formulation of query-focused summarization common in the
literature (\citealt{xu-lapata-2020-coarse,vig-etal-2022-exploring};
\emph{inter alia}). Let $q$~denote an information-seeking request
(e.g., ``What is so special about the pulmonary arteries and veins?''
in Figure~\ref{figure:example}) and $\mathcal{P}=\{p_1, \dots, p_n\}$
be a set of passages most relevant to~$q$ (see top left in
Figure~\ref{figure:example}). We adopt a two-step approach where~$k$
most relevant passages are first retrieved and possibly reranked based
on the query alone, and then fed to a secondary model which
synthesizes the passages into a final
summary~$\mathcal{S} = \{s_1, \dots, s_m\}$ consisting of
sentences~$s_i$. An \emph{attributed} summary further includes
supporting evidence. For the sake of simplicity, we assume this is
expressed by embedded in-line citations (see bottom left in
Figure~\ref{figure:example}).  Each~$s_i$ has a set of citations
$\mathcal{C}_i = \{c_{i,1}, \dots, c_{i,k}\}$ where $c_{i,j}$ is a
passage ID signifying that sentence~$i$ is citing said
passage. For the summary~(a) in Figure~\ref{figure:example},
$\mathcal{C}_1 = \{[1]\}$, and $\mathcal{C}_2 = \{[2]\}$.


In the following we first explain how to train a  sequence-to-sequence model to 
 generate citations, and then  describe the
proposed plan-based models.

\subsection{Sequence-to-Sequence Model}\label{subsec:seq2seq}

Datasets for query-focused summarization such as \aquamuse\
\citep{kulkarni2020aquamuse} or \textsc{ASQA}
\citep{stelmakh-etal-2022-asqa} typically consist of
queries~$\mathcal{Q}$, passages~$\mathcal{P}$, and target summaries
$\mathcal{S}$. A vanilla sequence-to-sequence model, generates
summary~$S$ given passages~$P$ and query~$q$, hence modeling the
conditional probability~$p(\mathcal{S|}P,q)$.  In the case of
attributed generation, the model is trained on summaries with
citations assumed to be part of the summary's token vocabulary. In
Figure~\ref{figure:example}, the query, passages, and summary~(a) with
in-line references to passages~[1] and~[5] would constitute a training
instance for this model.


In order to generate text with attribution, we must somehow annotate
target summaries with citations. In experiments, we obtain
silver-standard citations, by measuring the entailment between
passage~\mbox{$p_j \in \mathcal{P}$} as the premise and
sentence~$s_i \in \mathcal{S}$ as the hypothesis. We annotate $S_i$
with~$c_{i,j}$ for the passage~$p_j$ with the highest entailment score
(see Section~\ref{sec:experimental-setting} for more details).

\subsection{Blueprint Models with Attribution}
\label{sec:blueprint:attribution}

\citet{narayan2022conditional} formalize three types of blueprint
models, which are purportedly suited to different generation tasks
(e.g., long vs short output/input). We repurpose their end-to-end
variant as it is a straightforward extension of the standard approach
described above.  Our model encodes query~$q$ and passages
$\mathcal{P}$, and then generates, $\mathcal{B};\mathcal{S}$, the
concatenation of blueprint $\mathcal{B}$ and output
summary~$\mathcal{S}$ in one decoding step.  We define $\mathcal{B}$
as a set of questions $B=\{b_1, \dots, b_k\}$ that are answered
in~$\mathcal{S}$ and act as a proxy for content selection and
planning. This definition is more flexible (compared to
question-answer pairs), as we do not require answers to be extracted
from passages, and are not limited to a specific style of answers or
questions such as those manifested in SQuAD
\citep{rajpurkar-etal-2018-know}. Questions can be more general while
answers can be abstractive, and represent a variety of syntactic
constructs beyond noun phrases (e.g., verbs, clauses, even sentences).

We assume the model has access to training data consisting of queries,
passages, blueprints, and summaries (See Figure 1). Blueprints are not
generally available, we explain how we obtain these automatically in
the next section and define two blueprint models: an abstractive model
(Figure~\ref{figure:example}(b)) and an extractive model
(Figure~\ref{figure:example}(c)); these differ on whether the
question-based plan is generated or copied from the input.

\paragraph{Abstractive Model}  
We augment the training data with blueprints by generating questions
for each target summary. We adopt an overgenerate-and-rank strategy,
generating multiple questions to be filtered based on their
overlap with the summary.

A challenge with blueprint models is that the output may be answering
a mix of general and specific questions. To avoid generating only one
style of question, we employ two datasets originally developed for
question answering, namely SQuAD~\citep{rajpurkar-etal-2018-know} and
Natural Questions \citep{natural_questions}. QA datasets typically
consist of question/paragraph/answer triples but can be repurposed for
question generation; to produce general-purpose questions, we assume
the paragraph is the input and the question is the output, while for
more specific questions, we concatenate the paragraph with the
sentence where the answer span is found and the model's aim is to
generate a question for the specific answer.  We fine-tune
\mbox{T5-3B}~\citep{raffael_exploring_limits} to obtain these two
flavors of question generation models (see
Appendix~\ref{sec:qg-training} for training details).

For each question generation model, we sample (using beam search)
10~questions per sentence and another 10~per summary so as to have
a diverse question set.  For each summary sentence, we then select the
question with the highest lexical overlap from the union of questions (i.e.,~the questions
specific to that sentence \textit{and} the set of summary-level
questions). Blueprint questions are sorted according to the order of
appearance of their corresponding summary sentences.
We add a special character
in the blueprint (-- in Figure~\ref{figure:example}) to delineate the
questions corresponding to each sentence. We provide an example of how
questions are generated and subsequently filtered in
Appendix~\ref{sec:bp-selection}.


\paragraph{Extractive Model} In the extractive model, the questions align
more closely with the input passages. Specifically, we explicitly
verbalize the questions each passage might answer,
\mbox{i.e.,~$\mathcal{P}_{ext} = \{(p_1, \mathcal{Q}_1), \dots, (p_n,
\mathcal{Q}_n) \}$}, where~$p_i$ are retrieved passages and
$\mathcal{Q}_i$ corresponding questions.
Blueprint~$\mathcal{B}$ then consists of a subset of questions whose
answers are found in the input. The extractive model is functionally
equivalent to the abstractive one, the only difference being that it
encodes passages $\mathcal{P}_{ext}$ (instead of $\mathcal{P}$). This
model also outputs~$\mathcal{B}$;$\mathcal{S}$, however, the decoder
learns to \emph{copy} questions from the input in order to
construct~$\mathcal{B}$. The extractive model has two
advantages. Firstly, it reduces the risk of hallucinations and
irrelevant information since the questions are directly copied from
the input. Secondly, attribution comes for free, as we know which
passage gave rise to which questions (see
Figure~\ref{figure:example}). An obvious drawback is computational
efficiency, as questions must be generated for every input passage.

We create a large number of questions for each passage using the same
generation models described above. We then greedily select five
questions~$\mathcal{Q}_i$ for each~$p_i$ such that they have highest
lexical overlap with the passage and minimal overlap with each
other. Subsequently, we construct the extractive blueprints under the
assumption that their questions ought to be aligned to the input
passages \emph{and} output summary. We first discard questions that
cannot be answered in the target summary using an answerability
classifier, a T5 model trained on a repurposed version of SQuAD v2
\citep{rajpurkar-etal-2018-know} that predicts whether a question can
be answered by a passage (see Appendix~\ref{sec:answerability} for
training details).  The remaining questions are compared against the
target summary and for each sentence we select a question based on
lexical overlap.  As in the abstractive model, a special character is
added to mark which blueprint questions correspond to which sentences
(see Appendix~\ref{sec:bp-selection} for an example).

More formally, 
we summarize  the abstractive and extractive models  in Equations~\eqref{eq:abstractive} and \eqref{eq:extractive} respectively,
\begin{align}
    p(\mathcal{B};\mathcal{S} \, &| \, q, \{p_1, ..., p_n\})\label{eq:abstractive} \\
    p(\mathcal{B};\mathcal{S} \, &| \, q, \{(p_1, Q_1), ..., (p_n, Q_n)\}) \label{eq:extractive}
\end{align}
Note that \(Q_i\) is generated only using \(p_i\) so there is no dependency on the target data. While the target blueprint and its summary (\(\mathcal{B};\mathcal{S}\)) have been created based on the input, the model is trained to generate one after the other.

\section{\aquamuse{} Experiments}
\label{sec:aquamuse}

\subsection{Experimental Setting}
\label{sec:experimental-setting}



\paragraph{Data} The bulk of our experiments use \aquamuse{}
\citep{kulkarni2020aquamuse}, a long-form question answering dataset,
which simulates how a search engine might synthesize documents
of high relevance to a user query. It consists of queries taken from
Natural Questions~\citep{kwiatkowski2019natural}, passages from Common
Crawl, and multi-sentence summaries from Wikipedia. We use the splits
released in \citet{kulkarni2020aquamuse} which contain~6,599, 714, and
849 examples for training, development, and testing, respectively. The
average query/response length is 9.2/107.3 words, and the number of
input passages is 10.

All our models operate on the same passages which are reranked using T5-R \citep{huebscher2022zero}, a T5~11B-based encoder trained with a classification loss on the MS~MARCO dataset \citep{nguyen2016ms}. 
T5-R reorders passages $p \in \mathcal{P}$ based on  their relevance with query~$q$.
Target summaries in \aquamuse{} \citep{kulkarni2020aquamuse} were
annotated with in-line citations (see summaries in Figure~1) using
an entailment classifier (T5-11B; \citealt{raffael_exploring_limits}) fine-tuned on  the ANLI dataset
\citep{nie-etal-2020-adversarial}; neutral and contradicting pairs
were classified as non-entailing
\citep{honovich-etal-2022-true-evaluating}. We consider citations
 part of the summary's token vocabulary, but exclude them when 
evaluating summary quality. 



\paragraph{Comparison Models}\label{subsec:models_comp}
We compare several models, examining how attribution and
planning interact:
a sequence-to-sequence baseline which does not have a plan or perform
attribution~($-$Blueprint $-$Attribution); a model which generates
citations without a blueprint~($-$Blueprint $+$Attribution;
Section~\ref{subsec:seq2seq}); its mirror image, i.e.,~a model with a
blueprint (abstractive or extractive) but no attribution
($+$Blueprint$_{\mathcal{A}|\mathcal{E}}\hspace{.3ex}-$Attribution);
and a blueprint model with
attribution~($+$Blueprint$_{\mathcal{A}|\mathcal{E}}$
$+$Attribution). In all cases, we fine-tune a LongT5 model
(\citealt{guo-etal-2022-longt5}, 3B parameters).

\paragraph{Evaluation Metrics}\label{sec:evaluation}

We evaluate the quality of the summaries and for models that output
citations whether these are correct.  We measure the summary's
\emph{relevance} using ROUGE-L\footnote{\url{https://pypi.org/project/rouge-score/}} \citep{lin-2004-rouge} which computes
the longest common subsequence between output summaries and reference
text. We perform this comparison after removing any citations present
in the response.


We quantify the extent to which generated summaries are
\emph{faithful} to their input using textual entailment
\citep{maynez-etal-2020-faithfulness,falke-etal-2019-ranking,narayan-etal-2022-well,honovich-etal-2022-true}. Our
classifier is trained on the Adversarial NLI dataset
\citep{nie-etal-2020-adversarial}; for each sentence in the summary,
we check whether it is entailed by the input passages (and report the
average across all sentences to obtain an overall score). Again, we
remove citations from the output to measure faithfulness.

In addition, we evaluate \emph{citation quality} by checking whether
the citations mention the \emph{right} passages. We extend AutoAIS
\citep{bohnet2022attributed}, a recently proposed metric for
automatically measuring the attribution of short answers (in a QA
setting). For long-form responses, we take sentences with in-line
citations and check whether these are entailed by all the passages
they cite \citep{amplayo2023smart}.  We then report the proportion of
entailed sentences. We use the same ANLI classifier
\citep{nie-etal-2020-adversarial} described above.

For blueprint models, we would additionally expect the plan to convey
information relayed by the input passages. There is no point in
introducing an intermediate representation if it does not facilitate
grounding to the input. We measure \emph{blueprint quality}, by
leveraging an answerability classifier (see Appendix
\ref{sec:answerability} for details) that checks whether a question is
answered by a passage.  We report the proportion of questions in the
blueprint that are answered by the passages it references.\footnote{Note that we are not simply measuring whether the question is answerable by any input passage, but rather if it is answerable by the cited passage.} Questions
in the blueprint cite passages \textit{implicitly} through their
respective summary sentences. Recall that by construction there is a
one-to-one correspondence between blueprint questions and summary
sentences (which in turn cite input passages).

\begin{table}[t]
\resizebox{0.48\textwidth}{!}{
\centering
\footnotesize
  \begin{tabular}{llcc}
\toprule
  \multicolumn{2}{c}{Retrieval-based Models}  &   ROUGE-L & ANLI\\ 
    \midrule
    $-$Blueprint & $-$Attribution & 63.80 & 87.20  \\
        $+$Blueprint$_{\mathcal{A}}$ & $-$Attribution & 64.17 & 87.88 \\
    $+$Blueprint$_{\mathcal{E}}$ & $-$Attribution & 72.25 & 87.92 \\ \midrule
    $-$Blueprint & $+$Attribution & 63.72 & 86.73  \\
    $+$Blueprint$_{\mathcal{A}}$& $+$Attribution & 63.49 & \textbf{88.05} \\
   $+$Blueprint$_{\mathcal{E}}$& $+$Attribution & \textbf{72.98} & 88.01\\
  \bottomrule
  \end{tabular}
  }
  \caption{\aquamuse{} results: summary quality.}
  \label{tab:aquamuse-results}
\end{table}

\begin{table}[t]
\resizebox{0.48\textwidth}{!}{
  \begin{tabular}{llcc}
\toprule
  \multicolumn{2}{c}{Retrieval-based Models}  &   Answerability & AutoAIS\\ 
    \midrule
    $-$Blueprint & $+$Attribution & --- & 72.64  \\
     $+$Blueprint$_{\mathcal{A}}$& $+$Attribution & 92.27 & 74.16\\
     $+$Blueprint$_{\mathcal{E}}$& $+$Attribution & \textbf{97.97} &
    \textbf{74.35} \\
    \bottomrule
  \end{tabular}
  }
  \caption{\aquamuse{} results: attribution quality.}
  \label{tab:aquamuse:attribution}
\end{table}


\subsection{Results}
\label{sec:results:analysis}

\paragraph{The extractive blueprint performs best across evaluation metrics.}
Table~\ref{tab:aquamuse-results} presents our results on \aquamuse{}
when evaluating summary relevance and faithfulness. In the first block
we compare models that do not generate citations. As can be seen
plan-based models deliver better summaries in terms of ROUGE-L, with
slightly enhanced faithfulness (see ANLI column),
 with the extractive
blueprint model taking the lead. Our blueprint models are overall
better compared to \citet{narayan2022conditional}, achieving a new
state of the art in the \aquamuse{} dataset. However, as mentioned
earlier, the two approaches are not directly comparable as they differ
in terms of the input they expect and blueprint definition.  In the
second block of Table~\ref{tab:aquamuse-results} we compare models
with an attribution mechanism. In general, faithfulness increases for
plan-based models and relevance improves by 10~points
for the extractive blueprint model.

In Table~\ref{tab:aquamuse:attribution} we evaluate attribution
quality for models which output citations. We observe gains in AutoAIS
for our blueprint models over performing attribution without a plan.
Attribution quality is slightly higher for the extractive model, which
also creates more grounded blueprints, 97.97\%~of the questions are
answered by the passages it references. We show example output in
Appendix~\ref{sec:examples}.





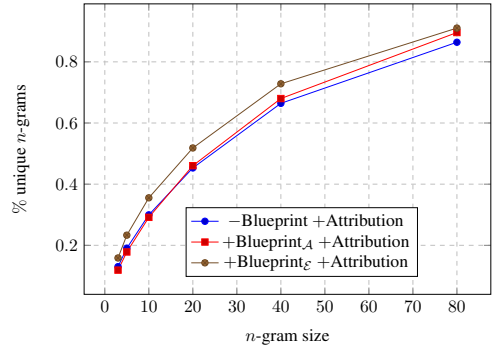
\begin{figure}[t]
  \centering
\begin{tikzpicture}[scale=.6]
	\begin{axis}[
		height=8cm,
		width=10.5cm,
		grid=major,
                grid style = dashed,
                legend style={at={(0.25,0.18)},anchor=west},
                 x label style={at={(axis description cs:0.5,-0.1)},anchor=north},
    y label style={at={(axis description cs:-0.12,.5)},anchor=south},
    xlabel={$n$-gram size},
    ylabel={$\%$ unique $n$-grams}]
	]
		
	\addlegendentry{$-$Blueprint~$+$Attribution}

	\addplot coordinates {
		(3,0.13027006 )
		(5,0.19068843)
		(10,0.29964921)
		(20,0.45302634)
		(40,0.66433342)
		(80,0.86410482)

	};
	\addlegendentry{$+$Blueprint$_{\mathcal{A}}$~$+$Attribution}
        \addplot coordinates {
		(3,0.11888636)
		(5,0.17825849)
		(10,0.29146482)
		(20,0.46046839)
		(40,0.6800103)
		(80,0.89587106)

	};

        	\addlegendentry{$+$Blueprint$_{\mathcal{E}}$~$+$Attribution}
        \addplot coordinates {
		(3,0.15861722 )
		(5,0.23319417)
		(10,0.35531461)
		(20,0.51839981)
		(40,0.7282454)
		(80,0.91044816)

	};
        
	\end{axis}
\end{tikzpicture}
\vspace*{-.1cm}
\caption{Unique $n$-grams in generated summary.}
\vspace{-20pt}
    \label{fig:abstractiveness}
\end{figure}

\paragraph{The extractive blueprint generates the most abstractive summaries.} We investigate the hypothesis that
blueprint models might perform better simply because it is easier for
them to copy sentences or passages from the input.  In
Figure~\ref{fig:abstractiveness} we evaluate the proportion of unique
$n$-grams in the generated summary (compared to the input passages)
for different $n$-gram sizes (where \mbox{$n=\{3,5,10,20,40,80\}$})
for the~$+$Attribution models in Table~\ref{tab:aquamuse-results}. As can
be seen, the extractive model generates the most abstractive
responses, while the $-$Blueprint baseline copies longer chunks of
text from the input. This suggests that blueprints help models
consolidate information from multiple sources, while still being
faithful to the input.

\begin{figure*}[t]
  \centering
  \small
  
\includegraphics[width=\textwidth]{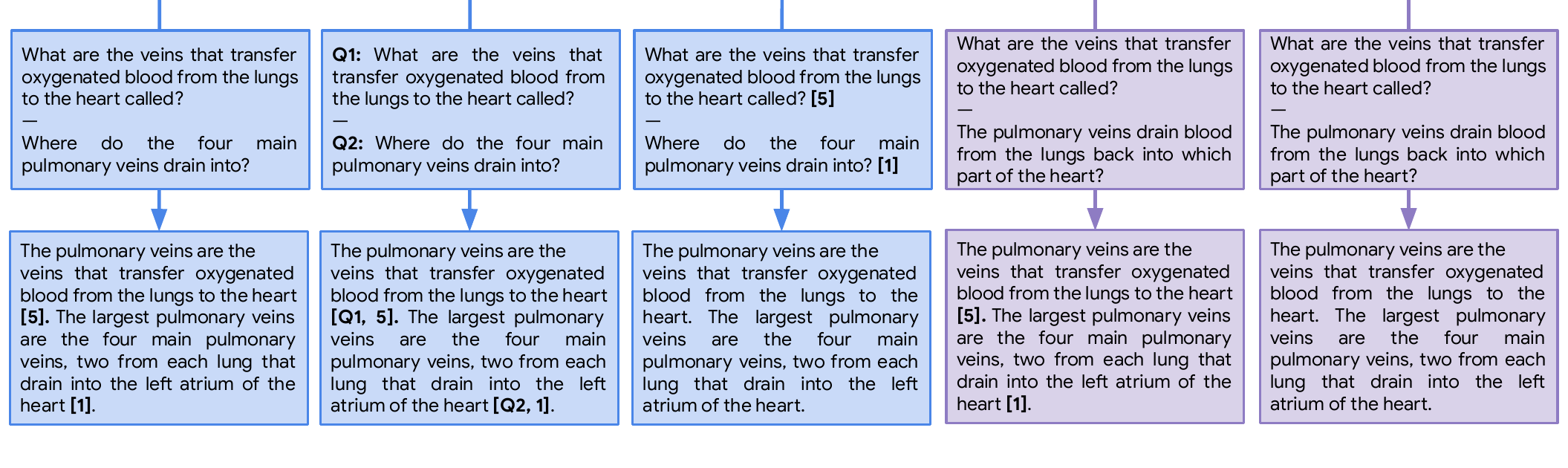}
\begin{tabular}{c@{\hspace{.6cm}}c@{\hspace{.4cm}}c@{\hspace{1cm}}c@{\hspace{1cm}}c}
    \raisebox{.3cm}[0pt]{In-line Citations} & \raisebox{.3cm}[0pt]{Question Citations} & \raisebox{.3cm}[0pt]{Blueprint Citations} &
    \raisebox{.3cm}[0pt]{In-line Citations} & \raisebox{.3cm}[0pt]{Implicit Citations} \\
 $b_1 \dots   b_k$ & $Q_1$: $b_1 \dots Q_k$:   $b_k$ & $b_1 $ [$c$]
    $\dots b_k$ [$c$] & $b_1 \dots b_k$ & $b_1 \dots b_k$\\
    $s_1$ [$c$]  $\dots s_m$ [$c$] & 
      $s_1$ [$Q_1, c$] $\dots$ $s_m$ [$Q_k, c$] &  $s_1 \dots s_m$ & $s_1$ [$c$]
    $\dots s_m$ [$c$] & $s_1 \dots s_m$ \\
  \end{tabular}
  
  \caption{\label{figure:formats} Blueprints (top), corresponding
    summaries (bottom), and different citation formats:
 $b_i$ is a blueprint question, $s_i$ is a summary sentence and $c,
    Q$ are citations.  Abstractive/extractive blueprints are colored
    in blue/purple.}
\end{figure*}


\paragraph{Blueprint plans  allow us to
define different citation formats.} Our experiments have so far
adopted in-line citations as a common format for all models. We now
explore alternatives which are unique to our plan-based models
and examine whether these have any bearing on
performance. Specifically, we (a)~augment in-line passage citations
with references to blueprint questions; (b)~place in-line citations in
the blueprint (after each question) rather than in the output summary;
and (c)~have implicit citations for extractive models only; in this
case, we do not generate any citations but attribution can be inferred
on the basis of the questions being aligned to summary sentences and
copied from the input passages. Figure~\ref{figure:formats} shows
examples and definitions of these formats.

Table~\ref{tab:format-ablation} reports results on plan quality
(checking whether the questions can be answered by the cited passages)
and citation correctness (AutoAIS). We see that formats assuming an
explicit connection between passages and questions improve
answerability, however, at the expense of AutoAIS. Overall, in-line
citations strike the best balance between answerability and
faithfulness. We further examine the quality of blueprint plans in
Appendix~\ref{sec:grounding}.

\paragraph{Human evaluation corroborates automatic attribution
  results.} In addition to automatic evaluation, we conducted a human
study to assess the attribution quality of our models. Participants
were asked (see Appendix~\ref{sec:instructions} for details on
our instructions and study) to verify whether the citations in the
generated summaries are correct, classifying each sentence as fully
supported (FSupp) by the referenced passages, partially supported
(PSupp), unsupported (NSupp), or contradictory (Contra). We elicited
judgments on 50~instances (randomly sampled from the AQuaMuse test
set) for the systems shown in Table~\ref{tab:humans} (first block). We
find that the extractive system is perceived as the most accurate
among attribution models achieving the highest proportion of correct
citations.


\begin{table}[t]
    \footnotesize
    \centering
    \begin{tabular}{@{}c@{\hspace*{1cm}}cc}
    \toprule
    {$+$Blueprint$_{\mathcal{A}}~+$Attribution}  &  Answer & AutoAIS \\\midrule
\noalign{\vskip 2pt}
    In-line Citations  & 92.27 & 74.16 \\
    Question Citations &  92.95 & 60.86 \\
    Blueprint  Citations & 93.67 & 61.49 \\
    \midrule
    $+$Blueprint$_{\mathcal{E}}~+$Attribution & Answer & AutoAIS
    \\ \midrule
\noalign{\vskip 2pt}
    In-line Citations  & 97.97 & 74.35 \\
    Implicit Citations & 98.20 & 63.79 \\
    \bottomrule
    \end{tabular}
    \caption{Results for different citation formats.}
    \label{tab:format-ablation}
\end{table}



\section{ALCE Experiments}

\subsection{Experimental Setting}
\paragraph{Data}
We further examine whether the attribution capabilities of our models
transfer to other domains, by performing (zero-shot) experiments on a
recently collated benchmark called ALCE \citep{gao2023enabling}.
ALCE  contains evaluation sets from ASQA
\citep{stelmakh-etal-2022-asqa}, which focuses on ambiguous queries
and long-form answers, ELI5 \citep{fan-etal-2019-eli5}, which focuses
on how/why/what questions that require explanations, and QAMPARI
\citep{rubin2022qampari} where the answers are lists of entities.
These datasets were selected such that the answers are factual,
long-form, and require information from multiple sources. In
experiments,  we use the passages provided by the ALCE benchmark with no
further reranking. Details on ALCE and representative
examples are in  Appendix~\ref{sec:datasets}.

\begin{table}[t]
  \resizebox{0.48\textwidth}{!}{
  \footnotesize
\begin{tabular}{@{}l@{\hspace*{-.5cm}}llclc@{}} \toprule
 \multicolumn{2}{c}{\aquamuse} &                    \multicolumn{1}{c}{\(\uparrow\)FSupp} &      \(\downarrow\)PSupp  &    \multicolumn{1}{c}{\(\downarrow\)NSupp}  &      \(\downarrow\)Contra\\\midrule
 $-$Blueprint & $+$Attribution  &           46.91$^{*}$   &      5.21         &        46.91$^{*}$           &            0.69 \\ 
 $+$Blueprint$_{\mathcal{A}}$ & $+$Attribution&       49.41$^{*}$   &      5.62    &             43.64          &             1.33\\
\textbf{$+$Blueprint$_{\mathcal{E}}$} & \textbf{$+$Attribution}  &       \textbf{58.28}    &     \textbf{4.10}      &           \textbf{36.90}
&              \textbf{0.71}\\ \midrule
 \multicolumn{2}{c}{ELI5 \& ASQA} &                    \multicolumn{1}{c}{\(\uparrow\)FSupp} &      \(\downarrow\)PSupp  &    \multicolumn{1}{c}{\(\downarrow\)NSupp}  &      \(\downarrow\)Contra\\\midrule
 Vicuna-13B & & 43.97$^{*}$ & 1.88 & 54.02$^{*}$ & 0.13 \\
 ChatGPT (Vanilla) & & 76.51$^{*}$ & 1.95 & 20.21$^{*}$ & 1.33 \\
 \textbf{$+$Blueprint$_{\mathcal{A}}$} &  \textbf{$+$Attribution}& \textbf{88.11}  & \textbf{3.50} & { }{ }\textbf{7.51} & \textbf{0.88}            \\\bottomrule
\end{tabular}
  }
  \caption{\label{tab:humans} Proportion of sentences with citations
    Fully Supported by the referenced passages, Partially Supported,
    Not Supported and in Contradiction. Systems marked with~$^{*}$ are
    significantly different the model in \textbf{bold}
    (\mbox{$p<0.01$}, using bootstrap resampling).}
\end{table}



\paragraph{Comparison Models} \citet{gao2023enabling} use ALCE to
examine the citation abilities of LLMs. They report experiments with
two base LLMs, namely ChatGPT (\texttt{gpt-3.5- turbo-0301}) and LLaMA
\citep{touvron2023llama}, both of which were prepended with detailed
instructions and a few demonstrations for in-context learning. Their
ChatGPT pipelines include (a) Vanilla, an end-to-end model where the
input is the query and top-5 passages; (b) Summarization condenses the
passages first with a separate LLM and thus can process more of them;
(c) Snippet extracts snippets from passages using an LLM; (d) Inline
Search, where instead of providing passages in the input, the LLM
calls search in-line whenever needed; and (e) Closed Book, where no
passages are provided and attribution happens post-hoc (i.e., for each
sentence, find the best passage to cite).  LLaMA-based pipelines are
versions of the Vanilla approach described above with either~(f)
LLaMA-13B, or (g)~Vicuna-13B \citep{vicuna2023}.\footnote{We did not include
 systems that additionally use response reranking and interaction
 strategies as these are orthogonal improvements and do not
 consistently improve performance.} 
 We compare our abstractive model
 ($+$Blueprint$_{\mathcal{A}}$\hspace*{.1ex}$+$Attribution) against
 these pipelines. We assume that a question generation model is not
 available at transfer time to obtain questions for input passages
 (and create blueprints for the extractive model).


\paragraph{Evaluation Metrics}\label{sec:evaluation-metrics} We follow
\citet{gao2023enabling} and evaluate system output in terms of
\emph{correctness} and \emph{attribution}. Correctness is measured
differently for each dataset; for ASQA, they calculate the recall of
correct short answers \cite{stelmakh-etal-2022-asqa}; for ELI5, they
use InstructGPT (\texttt{text-davinci-003};
\citealp{ouyang2022training}) to generate sub-claims from the target
response, and an NLI model (TRUE;
\citealp{honovich-etal-2022-true-evaluating}) to predict whether the
response entails each sub-claim; for QAMPARI, they calculate precision
and recall by checking the exact match to the gold-standard list of
answers, considering recall to be 100\% if the prediction includes at
least five correct answers \cite{rubin2022qampari}; here, we only use
recall to measure correctness since precision is calculated over a
list of predicted answers and our models output \emph{one} long answer
(given the \aquamuse{} training). Attribution is evaluated using
citation recall, which determines if the output is entirely supported
by cited passages, and citation precision, which identifies any
irrelevant citations.  We report F1 for metrics that have both
precision and recall.


\begin{table}[t]
\centering
\footnotesize
\resizebox{0.48\textwidth}{!}{
\begin{tabular}{llrrrrrrrr}
\toprule
& & \multicolumn{2}{c}{ASQA} & \multicolumn{2}{c}{ELI5} & \multicolumn{2}{c}{QAMPARI} & \multicolumn{2}{c}{Average} \\
 & & \multicolumn{1}{c}{C} &  \multicolumn{1}{c}{A} &  \multicolumn{1}{c}{C} &  \multicolumn{1}{c}{A} &  \multicolumn{1}{c}{C} &  \multicolumn{1}{c}{A} &  \multicolumn{1}{c}{C} &  \multicolumn{1}{c}{A} \\
\midrule
\noalign{\vskip 2pt}
& Vanilla (5-psg) & 40.4 & 73.0 & 12.0 & 50.5 & 20.8 & 20.7 & 24.4 & 48.1 \\
& Summarization (10-psg) & \textbf{43.3} & 65.2 & 12.3 & 49.8 & 22.3 & \textbf{24.6} & 26.0 & 46.5 \\
& Snippet (10-psg) & 41.4 & 61.1 & 14.3 & 47.5 & 22.9 & 23.9 & 26.2 & 44.2 \\
& Inline Search & 32.4 & 58.2 & \textbf{18.6} & 44.6 & 18.7 & 14.9 & 23.2 & 39.3 \\
\raisebox{1ex}[0pt]{\begin{sideways}{ChatGPT}\end{sideways}}& Closed Book & 38.3 & 26.7 & \textbf{18.6} & 15.5 & \textbf{24.7} & 10.0 & \textbf{27.2} & 17.4 \\
\midrule
& LLaMA-13B & 26.9 & 12.6 & 3.9 & 3.9 & 9.4 & 6.9 & 13.4 & 7.8 \\
& Vicuna-13B & 31.9 & 50.6 & 10.0 & 17.4 & 14.9 & 12.9 & 18.9 & 27.0 \\
\midrule
& LongT5~3B (10-psg) \\
& \hspace*{-.25cm}$+$Blueprint$_{\mathcal{A}}$ $+$Attribution  & \raisebox{.18cm}[0pt]{33.8} & \raisebox{.18cm}[0pt]{\textbf{77.8}} & \raisebox{.18cm}[0pt]{5.2} & \raisebox{.18cm}[0pt]{\textbf{60.9}} & \raisebox{.18cm}[0pt]{12.9} & \raisebox{.18cm}[0pt]{10.8} & \raisebox{.18cm}[0pt]{17.3} & \raisebox{.18cm}[0pt]{\textbf{49.8}} \\
\bottomrule
\end{tabular}
}
\caption{Results on ALCE benchmark:  LLM pipelines
 and proposed abstractive blueprint model.}
\label{tab:alce-results}
\end{table}

\subsection{Results}

\paragraph{The abstractive blueprint competes with LLMs
in attribution.} Our results are summarized in
Table~\ref{tab:alce-results}. On average, we observe that our model is
better than competing systems in terms of attribution.  We suspect
this is due to the fact that it has been explicitly fine-tuned with
outputs that have citations, which in turn improves attribution,
rendering the model robust across datasets. In terms of correctness,
our model performs on average better than LLaMA-13B but worse than the
other systems, which is not surprising as it is applied to new
datasets without access to in-domain training. In contrast, comparison
LLM pipelines have been exposed to few-shot demonstrations.

\paragraph{Humans find Blueprint citations more accurate compared to those produced by LLMs.} To further
validate the results in Table~\ref{tab:alce-results}, we conducted
human evaluation comparing ChatGPT Vanilla, Vicuna-13B, and our
abstractive blueprint model.
Using the same instructions from Section~\ref{sec:results:analysis},
we collected judgments over 100 randomly sampled  instances (50 from 
ELI5 and 50 from ASQA).\footnote{For  
  \citeauthor{gao-2022-systeme}'s \shortcite{gao2023enabling} systems, we use the output  from
  \url{https://github.com/princeton-nlp/ALCE/tree/main/human_eval}.  Appendix~\ref{sec:alce-human-detailed} shows results for each dataset.} As shown in Table~\ref{tab:humans}, 
our model is perceived to have the most accurate
citations, significantly outperforming both Vicuna-13B and ChatGPT Vanilla.


\section{Conclusion}

In this paper we focused on retrieval augmented generation with
citations. We explored the attribution capabilities of plan-based
models and proposed different variants depending on how the plans are
created (i.e.,~abstractively or extractively). Our experiments
revealed several interesting findings. Summary quality improves for
blueprint models when these learn to generate citations, both in terms
of relevance and faithfulness. Conversely, this is not the case for
models without an intermediate planning stage, in fact attribution
slightly hurts performance (see second line in Table
\ref{tab:aquamuse-results}). The attribution mechanism is also
important; formulating the attribution as in-line citations seems
beneficial as well as a tight alignment between the input and the
blueprint questions. Finally, attribution is a transferable skill for
blueprint models, across datasets and information-seeking tasks.


\section*{Limitations}

In this work, we have focused exclusively on encoder-decoder
models. Future work could explore fine-tuning (on attribution
annotations) a wider range of models, including decoder-only ones. In
the future we would further like to explore more efficient ways of
indexing passages with questions, as well as using the questions as an
additional training signal for improving information
retrieval. Finally, an important consideration with generative models
is the problem of misinformation. While the work we present here makes
a step towards improving the faithfulness and factual consistency of
summarization systems, it is important to note that our models still
make mistakes, in particular in out-of-domain scenarios.

\section*{Acknowledgments} 
We thank the anonymous reviewers for their feedback. We are grateful to Hannah Rashkin, Chris Alberti, Slav Petrov, and Dipanjan Das for their comments on earlier versions of this paper. 

\bibliography{anthology,custom}
\bibliographystyle{acl_natbib}

\newpage

\appendix

\section{Question Generation Implementation}
\label{sec:qg-training}

Two question generation models were used to obtain questions for input
passages and blueprints: one model was trained on SQuAD v2
\citep{rajpurkar-etal-2018-know} and another one on Natural Questions
\citep{natural_questions}. Both question generators were built on top
of a T5~3B \citep{raffael_exploring_limits} base model, and expect
either of the following input formats:
\begin{enumerate}
    \item General-Purpose QG, where the model generates a question
      that can be answered anywhere in the passage:
    
    \begin{quote}
        \texttt{Generate question >>> [PASSAGE]}
    \end{quote}
    
    \item Sentence-Specific QG, where the model generates a question
      specific to a sentence in the passage:
    
    \begin{quote}
        \texttt{Generate question >>> [PASSAGE] >> [SENTENCE]}
    \end{quote}
\end{enumerate}
where \texttt{[PASSAGE]} and \texttt{[SENTENCE]} are the passage and
the sentence where the answer to the question (to-be generated) is found.

\section{Answerability Classifier Implementation}
\label{sec:answerability}

Our answerability classifier repurposes the SQuAD~v2
\citep{rajpurkar-etal-2018-know} dataset that includes unanswerable
questions.  Specifically, a T5~11B model is fine-tuned to accept as
input a question and a passage in the following format:
\begin{quote}
\texttt{question: [question] context: [passage]}
\end{quote}
and to generate \texttt{Yes} as output in cases where
\texttt{[question]} can be answered given \texttt{[passage]} as
context and \texttt{No} otherwise.

\section{Blueprint Selection Examples}
\label{sec:bp-selection}

\subsection{Abstractive Blueprints}

Table~\ref{tab:abs-blp-selection} shows a small-scale example of how
questions are selected in abstractive blueprints.  Given the summary
in the example, we generate three SQuAD-style questions (marked with
subscript~$Sq$) and three NQ-style questions (marked with
subscript~$N$) for each summary sentence. In addition, we generate
three SQuAD and three NQ summary-level questions, giving us a total of
24 questions.  Note that in our experiments, we use 10 SQuAD/NQ
questions instead of three.  For each sentence, we then select the
question with the highest lexical overlap from the union of the
sentence-specific and summary-level questions. In our example, the
question for~$S_1$ is selected from the summary-level set. Finally, we
collate these questions together with diacritic~-- to create the
blueprint.

\begin{table*}[th!]
 \footnotesize
\center
 \resizebox{.95\textwidth}{!}{
\begin{tabular}{| p{16cm} |c|} \hline
\multicolumn{2}{|c|}{{Questions from Summary Sentences}}  \\ \hline
Sentence 1 Questions & \\
Q$_{Sq,1}$: What is the purpose of pulmonary veins? &  $S_1$\\
Q$_{Sq,2}$: What are the pulmonary veins? & \\
Q$_{Sq,3}$: What veins transfer oxygenated blood from the lungs to the heart? & \\
Q$_{N,1}$: where does the pulmonary vein carry blood to & \\
Q$_{N,2}$: where does the blood in the pulmonary vein come from & \\
Q$_{N,3}$: where do the pulmonary veins carry blood to & \\
\hdashline
Sentence 2 Questions & \\
Q$_{Sq,1}$: How many pulmonary veins are there? & \\
Q$_{Sq,2}$: How many pulmonary veins are in each lung? & \\
Q$_{Sq,3}$: Where do the four main pulmonary veins drain into? & $S_2$\\
Q$_{N,1}$: where does the blood in the pulmonary vein go & \\
Q$_{N,2}$: where does the pulmonary vein carry blood to & \\
Q$_{N,3}$: where do the pulmonary veins connect to the heart & \\
\hdashline
Sentence 3 Questions & \\
Q$_{Sq,1}$: What are pulmonary veins a part of? & \\
Q$_{Sq,2}$: The pulmonary veins are a part of what? & \\
Q$_{Sq,3}$: What circulation are the pulmonary veins a part of? & $S_3$ \\
Q$_{N,1}$: where does the pulmonary vein carry blood to & \\
Q$_{N,2}$: where do the pulmonary veins carry blood to & \\
Q$_{N,3}$: where does the blood in the pulmonary vein go & \\
\hdashline
Summary-Level Questions & \\
Q$_{Sq,1}$: What are the veins that transfer oxygenated blood from the lungs to the heart called? &  \\
Q$_{Sq,2}$: What are the veins that transfer oxygenated blood from the lungs to the heart? & \\
Q$_{Sq,3}$: How many main pulmonary veins are there? & $S$\\
Q$_{N,1}$: where does the pulmonary vein carry blood to & \\
Q$_{N,2}$: where does the blood in the pulmonary vein go & \\
Q$_{N,3}$: where do the pulmonary veins carry blood to & \\
\hdashline
\hline
\multicolumn{2}{|p{16cm}|}{Summary} \\ 
\multicolumn{2}{|p{16cm}|}{[The pulmonary veins are the veins that transfer oxygenated blood from the lungs to the heart.]$_{S_1}$ [The largest pulmonary veins are the four main pulmonary veins, two from each lung that drain into the left atrium of the heart.]$_{S_2}$ [The pulmonary veins are part of the pulmonary circulation.]$_{S_3}$}
\\
\hline
\end{tabular}
}
\caption{\label{tab:abs-blp-selection} Abstractive blueprint for an
  example summary. We split the summary into sentences, and generate a
  SQuAD-style question for each (see blocks $S_1$--$S_3$). We also create general-purpose, \mbox{summary-level} questions
  (see fourth block). We then select one question per summary sentence
  from the set of \emph{all} generated questions.}
\end{table*}

\subsection{Extractive Blueprints}

Table~\ref{tab:ext-blp-selection} shows a small-scale example of how
questions are selected in extractive blueprints.  For each of the five
input passages (10 in our experiments), we generate the same amount of
questions using the method described above.  That is, for each passage
we generate sentence-level and summary-level questions (similar to
what is shown in Table~\ref{tab:abs-blp-selection}) using both the
SQuAD and the NQ question generation models.  We then greedily select
five questions for each passage, such that they have the highest
lexical overlap with the passage and minimal overlap with each
other. Once we have the passage questions, we are ready to select the
questions in the blueprint.  We first filter out questions that cannot
be answered using an answerability classifier (see
Appendix~\ref{sec:answerability} and AF column in the table).  We then
select, for each sentence in the summary, the question from with the
highest lexical overlap(SS column in the table).  Finally, we collated
these questions together with diacritic~-- to create the blueprint.

\begin{table*}[th!]
 \footnotesize
\center
\resizebox{.95\textwidth}{!}{
\begin{tabular}{| p{16cm} |c|c|} \hline
\multicolumn{1}{|c|}{{Questions from Input Passages}} & AF & SS \\ \hline
Passage 1 Questions & &\\
Q$_1$: what is essential for bringing oxygenated blood from the lungs back to the heart & \cmark & \\
Q$_2$: the pulmonary veins drain blood from the lungs back into which part of the heart & \cmark & $S_2$ \\
Q$_3$: the pulmonary arteries connect the heart to what &  & \\
Q$_4$: what connects the heart to the lungs and allows the blood to get oxygenated & & \\
Q$_5$: what drains blood from the lungs back into the left atrium of the heart & \cmark & \\
\hdashline
Passage 2 Questions & &\\
Q$_1$: in the heart-lung circulation, what brings deoxygenated blood from the right side of the heart to the lungs to be oxygenated  & &\\
Q$_2$: where does the pulmonary artery bring blood back to & & \\
Q$_3$: how is oxygenated blood drained back to the heart & \cmark & \\
\hdashline
Passage 3 Questions & &\\
Q$_1$: what happens to oxygenated blood when it is pumped through arteries to other parts of the body & &\\
Q$_2$: what holds the secret to appropriate functioning of the circulatory and cardiovascular system & \cmark & \\
Q$_3$: what is the function of pulmonary veins in the heart & \cmark & \\
Q$_4$: the body can not live and grow without what & \cmark & \\
Q$_5$: pulmonary veins carry blood from which part of the body & \cmark & \\
\hdashline
Passage 4 Questions & &\\
Q$_1$: which pulmonary veins pass behind the right atrium and superior vena cava; the left in front of the descending thoracic aorta & & \\
Q$_2$: what are large blood vessels that carry oxygenated blood from the lungs to the left atrium of the heart & \cmark & \\
Q$_3$: the number of pulmonary veins opening into the left atrium can vary between three and how many in the healthy population & \cmark & \\
Q$_4$: the superior pulmonary vein lies in front of and a little below which artery & &\\
Q$_5$: what is the anterior surface of the bronchus invested by & &\\
\hdashline
Passage 5 Questions & &\\
Q$_1$: what are the veins that transfer oxygenated blood from the lungs to the heart called & \cmark & $S_1$ \\
Q$_2$: what is a main function of the pulmonary veins & \cmark & \\
Q$_3$: how many primary pulmonary veins does each lung have & \cmark & \\
Q$_4$: where do the pulmonary veins drain into the heart & \cmark & \\
Q$_5$: what circulation are the pulmonary veins a part of & \cmark & $S_3$ \\
\hline
\multicolumn{3}{|p{16cm}|}{Summary} \\ 
\multicolumn{3}{|p{16cm}|}{[The pulmonary veins are the veins that transfer oxygenated blood from the lungs to the heart.]$_{S_1}$ [The largest pulmonary veins are the four main pulmonary veins, two from each lung that drain into the left atrium of the heart.]$_{S_2}$ [The pulmonary veins are part of the pulmonary circulation.]$_{S_3}$}
\\
\hline
\end{tabular}
}
\caption{\label{tab:ext-blp-selection} Extractive blueprint 
  for an example summary. Each passage has at most five questions. We
  first filter out questions that cannot be answered based on the
  summary, using an answerability classifier (see AF column). We then
  select at most one question for each summary sentence based on 
  lexical overlap (see SS~column).}
\end{table*}

\begin{table*}[t]
  \centering
  \small
\begin{tabular}{lllcc} \\\toprule
  \multicolumn{1}{c}{Dataset}  & \multicolumn{1}{c}{Question Type} &  \multicolumn{1}{c}{\#passages}  &   \multicolumn{1}{c}{Question Length} & \multicolumn{1}{c}{Answer Length} \\\midrule
  ASQA &  Factoid (ambiguous) &  Wikipedia (21M) &12.5 & 65.7 \\
  QAMPARI &  Factoid (list) & Wikipedia (21M) & 10.5 & 13.0\\
  ELI5 &   Why/How/What & Sphere (899M) &  17.4 & \hspace*{-.1cm}120.6 \\ \bottomrule
\end{tabular}
\caption{\label{tab:alce:statistics} Datasets in the ALCE benchmark
  \citep{gao2023enabling}. Question and Answer Length refer to the
  average number of words; for QAMPARI, Answer Length is the average
  number of answers. Examples are shown in
  Table~\ref{tab:alce:examples}.}
\end{table*}

\section{Grounding and Controllability}
\label{sec:grounding}


We further examine whether the \textit{planning} aspect of the
blueprint models remains intact when they are also expected to
generate summaries with references to the input. We measure the extent
to which the output is grounded to the blueprint plan using the
answerability classifier. Specifically, we check if the questions in
the blueprint are answerable by the summary as a whole and by individual
sentences (and report the average). Table~\ref{tab:grounding} shows
our results for blueprint models with and without attribution.  As can
be seen, extractive blueprint models are less grounded compared to
their abstractive counterparts, even though the former are better at
attribution (see Table~\ref{tab:aquamuse:attribution}).
 
\begin{table}[t]
\footnotesize
    \centering
\vspace{-12pt}
   \begin{tabular}{@{}l@{~}l@{~}c@{~}c@{}}
    \toprule
     \multicolumn{2}{c}{Retrieval-based Models} & Summary & Sentences \\
    \midrule
    $+$Blueprint$_{\mathcal{A}}$ & $-$~Attribution & 96.67 & 96.49 \\
        $+$Blueprint$_{\mathcal{E}}$ & $-$~Attribution & 91.32 & 80.06 \\ 
 $+$Blueprint$_{\mathcal{A}}$ & $+$~Attribution    & 97.46 & 93.64 \\
$+$Blueprint$_{\mathcal{E}}$ & $+$~Attribution & 91.22 & 79.97 \\
    \bottomrule
    \end{tabular}
    
    \caption{Grounding results for Blueprint Models (with and without attribution). }
    \label{tab:grounding}
\end{table}

We further illustrate how to improve the attribution quality of
abstractive models, thereby showcasing the controllability aspect of
blueprint models. We simply remove (post-hoc) generated blueprint
questions that are unanswerable based on the input passages, and then
generate the summary using the filtered blueprint. This
post-processing substantially improves attribution by~7.09 AutoAIS
points (compared to Table~\ref{tab:aquamuse:attribution}) while
retaining or marginally improving output quality on other dimensions
(63.12 ROUGE-L, 89.45 ANLI, 93.08, Answerability, and 81.25 AutoAIS).


\section{Experimental Instructions}
\label{sec:instructions}

\begin{figure*}
    \centering
    \includegraphics[width=\textwidth]{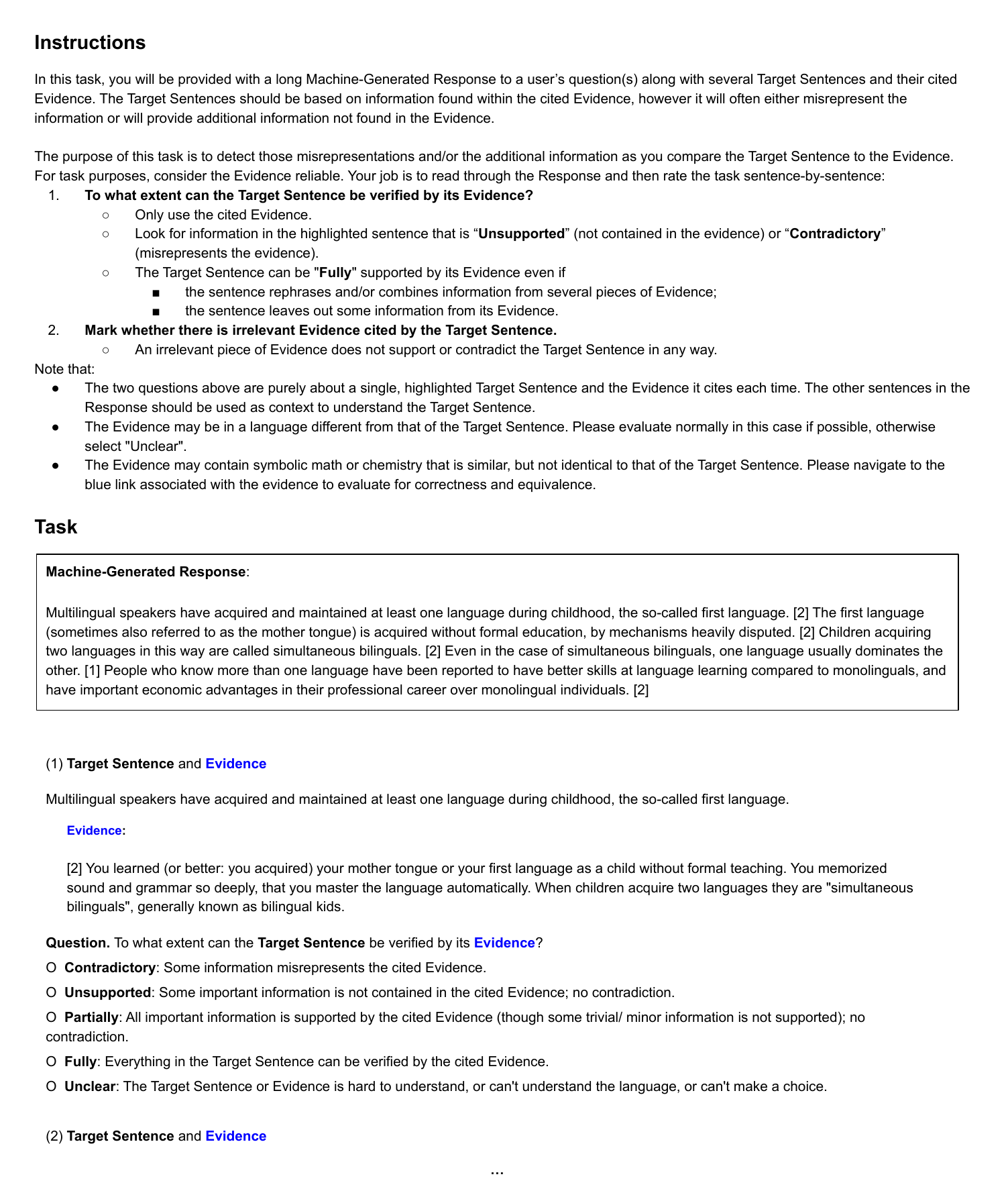}
    \caption{Experimental instructions presented to participants during the human elicitation study. The question repeats for each sentence in the machine-generated response.}
    \label{fig:human-eval-instr}
\end{figure*}

Figure~\ref{fig:human-eval-instr} presents the experimental instructions used in our human elicitation study. 
To ensure high quality annotations, we created a screener test to determine the raters’ suitability for the task.
We also limited the maximum tasks per annotator per study to six to avoid annotation fatigue.
A total of 108 annotators participated in the elicitation study for \aquamuse, and a total of 88 and 249 annotators participated in the studies for ASQA and ELI5, respectively.
The annotator agreement scores are as follows: 69.87\% on \aquamuse, 93.00\% on ASQA, and 89.04\% on ELI5.
Our annotators were paid adequately by our suppliers adhering to the supplier code of conduct.

\begin{table}[t]
  \resizebox{0.48\textwidth}{!}{
  \footnotesize
\begin{tabular}{@{}l@{\hspace*{-.5cm}}llclc@{}} \toprule
 \multicolumn{2}{c}{ASQA} &                    \multicolumn{1}{c}{FSupp} &      PSupp  &    \multicolumn{1}{c}{NSupp}  &      Contra\\\midrule
Vicuna-13B & & 57.87$^{*}$ & 3.40 & 38.30$^{*}$ & 0.43 \\
 ChatGPT (Vanilla) & & 78.02$^{*}$ & 1.61 & 18.77$^{*}$ & 1.61 \\
\textbf{$+$Blueprint$_{\mathcal{A}}$} &  \textbf{$+$Attribution}&
\textbf{88.58}& \textbf{2.75} & { }{ }\textbf{7.40} & \textbf{1.27}            \\ \midrule
 \multicolumn{2}{c}{ELI5} &                    \multicolumn{1}{c}{FSupp} &      PSupp  &    \multicolumn{1}{c}{NSupp}  &      Contra\\\midrule
 Vicuna-13B & & 37.57$^{*}$ & 1.17 & 61.25$^{*}$ & 0.00$^{*}$ \\
 ChatGPT (Vanilla) & & 75.76$^{*}$ & 2.12 & 20.93$^{*}$ & 1.19 \\
\textbf{ $+$Blueprint$_{\mathcal{A}}$} &  \textbf{$+$Attribution}& \textbf{87.42}  & \textbf{4.60} & { }{ }\textbf{7.67} & \textbf{0.31}            \\\bottomrule
 \end{tabular}
  }
  \caption{\label{tab:asqa-eli5-humans} Proportion of sentences with
    citations Fully Supported by the referenced passages, Partially
    Supported, Not Supported, and in Contradiction. Systems marked
    with~$^{*}$ are significantly different from the model in
    \textbf{bold} (\mbox{$p<0.01$}, using bootstrap resampling).}
\end{table}

\section{Human Evaluation Results on \textsc{ASQA} and ELI5}
\label{sec:alce-human-detailed}

Table~\ref{tab:asqa-eli5-humans} reports our human evaluation results
for \textsc{ASQA} and ELI5, respectively. Results mostly align with
the combined results in Table~\ref{tab:humans}, showing that our model
significantly produces more accurate citations.

\section{The ALCE Benchmark}
\label{sec:datasets}

ALCE \citep{gao2023enabling} is a collection of datasets aimed at
evaluating LLM citation capabilities. They contain factual questions
which require long-form answers aggregating information over multiple
sources (i.e.,~100-word passages). We summarize various statistics on
these datasets in Table~\ref{tab:alce:statistics} and describe them
below.

\begin{table*}
\small
  \centering
\begin{tabular}{lp{13cm}}  \\ \toprule
\multicolumn{2}{c}{ASQA} \\ \midrule
  Q: &  Who played  Bill
  Weasley in Harry Potter and the Prisoner of Azkaban?\\
 A: & The Weasely family is a family of wizards including Molly
  and Arthur Weasely as well as their seven children. Richard Fish
  appeared as Bill briefly in the film adaptation of Harry Potter and
  the Prisoner of Azkaban. Domhnall Gleeson, the son of actor Brendan
  Gleeson (Alastor Moody in the series), played Bill Weasley in Harry
  Potter and the Deathly Hallows. Chris Rankin played Percy Weasely
  and Fred and George Weasely were played by identical twins James
  Phelps and Oliver Phelps. Rupert Grint played Ron
  Weasely. \\ \bottomrule
  \multicolumn{2}{c}{QAMPARI} \\ \midrule
  Q: & Who won a Coke Zero Sugar 400 competition? \\
  A: & Sam McQuagg, Bobby Allison, Fireball roberts, Jimmie Johnson,
  Tony Stewart, Tim Richmond, Jeff Gordon, David Ragan, Erik Jones,
  Tony Stewart, Jamie McMurray, Kyle Busch, Cale Yarborough, Dale
  Earnhardt Jr., Brad Keselowski, David Pearson, Dale Jarrett, Jeff
  Gordon, Tony Stewart, Greg Biffle, A. J. Foyt, Richard Petty, Justin
  Haley, Tony Stewart. \\ \bottomrule
  \multicolumn{2}{c}{ELI5} \\ \midrule
    Q: & How does so much of our trash end up in the ocean?\\
    A: & Because water flows downhill and very often ends up in rivers
    which very often end up in oceans. So when it rains, trash is
    washed downhill and into streams and rivers and ultimately the
    ocean. \\ \bottomrule
\end{tabular}
\caption{\label{tab:alce:examples} Examples of question-answer pairs
  from ALCE datasets; input passages are omitted for the sake of brevity.}
\end{table*}

\paragraph{ASQA}\hspace*{-.3cm}\citep{stelmakh-etal-2022-asqa} focuses
on ambiguous questions which have multiple interpretations (see
Table~\ref{tab:alce:examples} for an example). The questions were
taken from AmbigQA \citep{min-etal-2020-ambigqa}, while long-form
answers were crowdsourced by synthesizing information from multiple
documents. For ALCE \citep{gao2023enabling}, ASQA questions were
additionally paired with Wikipedia passages
(\mbox{2018-12-20}~snapshot) which purportedly contained the answers.

\paragraph{QAMPARI}\hspace*{-.3cm}\citep{rubin2022qampari} is an
open-domain QA dataset where answers are lists of entities, drawn from
different passages.  All questions in QAMPARI have at least 5~answers,
with an average of 13~answers. Multi-answer questions were
automatically generated using manually defined templates, answers were
collated from Wikipedia, and examples were verified and paraphrased by
crowdworkers (see Table~\ref{tab:alce:examples} for an example). For
ALCE \citep{gao2023enabling}, QAMPARI questions were also paired with
Wikipedia passages (\mbox{2018-12-20}~snapshot).

\paragraph{ELI5}\hspace*{-.3cm}\citep{fan-etal-2019-eli5} mostly
constists of how/why/what questions that require in-depth long answers
and multiple passages as evidence. Questions (and answers) were
elicited from the subreddit \emph{Explain Like I’m Five} (ELI5) where
users are encouraged to provide answers which are comprehensible by a
five year old (see Table~\ref{tab:alce:examples}). For ALCE
\citep{gao2023enabling}, ELI5 questions were paired with passages from
Sphere \citep{Piktus:ea:2021}, a filtered version of Common Crawl.

ALCE \citep{gao2023enabling} contains 1,000 randomly selected examples
from each dataset (development set). It does not provide training data
as it is aimed at assessing the citation capabilities of LLMs.




\section{Example Output}
\label{sec:examples}

Tables~\ref{tab:output:example:1}, \ref{tab:output:example:2}, and~\ref{tab:output:example:3} present  output examples for the baseline system
($-$Blueprint~$+$Attribution) and our two blueprint variants ($+$Blueprint$_{\mathcal{A|E}}$~$+$Attribution). Blueprints created by the  plan-based systems are respectively shown in Tables~\ref{tab:blueprints:1}, \ref{tab:blueprints:2}, and~\ref{tab:blueprints:3}.  In general, we observe that the baseline system considers a smaller number of input passages when generating its response, while plan-based summaries contain more diverse passage references. As far as the blueprints are concerned, we find that the abstractive system generates (on average) more questions compared to the extractive model and the reference.  Although the questions making up the blueprints of the two models are not identical, we see that they cover similar topics. For instance, in Table~\ref{tab:blueprints:1} the blueprints emphasize topics like the mean temperatures of a tropical monsoon climate, its precipitation, dry seasons, and variance in temperature.

\begin{table*}[t]
\small
\center
\begin{tabular}{lp{13cm}} 
\multicolumn{2}{c}{describe the seasonal patterns of the tropical savanna biome} \\ \toprule
$[$0$]$ & However, unlike most tropical climates, a sizeable amount of precipitation falls during the dry season. In essence, this version of the monsoon climate generally has less pronounced dry seasons than tropical savanna climates $\dots$ \\
$[$1$]$ &Tropical monsoon climates have monthly mean temperatures above 18~$^{\circ}$C in every month of the year and feature wet and dry seasons, as Tropical savanna climates do. Tropical monsoon climates however features its driest month seeing on average less than 60~mm and this latter fact is in direct contrast to a tropical savanna climate $\dots$\\
$[$2$]$ & 64. Tropical wet and dry or savanna climate (Aw) have a pronounced dry season, with the driest month having precipitation less than 60 mm and less than 1/25 of the total annual precipitation. Example(s) of areas with this climate is(are) $\dots$\\
$[$3$]$ & Tropical monsoon climate, occasionally also known as a tropical wet climate or tropical monsoon and trade-wind littoral climate in climate classification, is a relatively rare type of climate. Tropical monsoon climates have monthly mean temperatures above 18~$^{\circ}$C (64~$^{\circ}$FF) $\dots$\\
$[$4$]$ & The second letter indicates the precipitation pattern—w indicates dry winters (driest winter month average precipitation less than one-tenth wettest summer month average precipitation. s indicates at least three times as much rain in the wettest month of winter as in the driest month of summer  $\dots$\\ \toprule
\multicolumn{2}{c}{$-$Blueprint $+$Attribution} \\ \toprule
& Tropical monsoon climate, occasionally also known as a tropical wet climate or tropical monsoon and trade-wind littoral climate in climate classification, is a relatively rare type of climate. [3] Tropical monsoon climates have monthly mean temperatures above 18~$^{\circ}$C (64~$^{\circ}$F) in every month of the year and feature wet and dry seasons, as Tropical savanna climates do. [3] Unlike tropical savanna climates however, a tropical monsoon climate's driest month sees less than 60 mm (2.36 in) of precipitation. [3] Also a tropical monsoon climate tends to see less variance in temperatures during the course of the year than a tropical savanna climate. [3] This climate has a driest month which nearly always occurs at or soon after the winter solstice for that side of the equator. [3]\\ 
\bottomrule
\multicolumn{2}{c}{$+$Blueprint$_{\mathcal{A}}$ $+$Attribution} \\ \toprule
& Tropical monsoon climate, occasionally also known as a tropical wet climate or tropical monsoon and trade-wind littoral climate in climate classification, is a relatively rare type of climate. [3] Tropical monsoon climates have monthly mean temperatures above 18~$^{\circ}$C (64~$^{\circ}$F) in every month of the year and feature wet and dry seasons, as tropical savanna climates do. [3] Unlike tropical savanna climates however, a tropical monsoon climate's driest month sees less than 60 mm (2.36 in) of precipitation. [3] Additionally, a tropical monsoon climate tends to see less variance in temperatures during the course of the year than a tropical savanna climate. [1] This climate has a driest month which nearly always occurs at or soon after the "winter" solstice for that side of the equator. [3] There are generally two versions of a tropical monsoon climate: Less pronounced dry seasons and Extraordinarily rainy wet seasons and pronounced dry seasons. [0] Regions with this variation of the monsoon climate typically see copious amounts of rain during the wet season. [1]\\ \bottomrule
\multicolumn{2}{c}{$+$Blueprint$_{\mathcal{E}}$ $+$Attribution} \\ \toprule
& Tropical monsoon climate, occasionally also known as a tropical wet climate or tropical monsoon and trade-wind littoral climate in climate classification, is a relatively rare type of climate. [3] Tropical monsoon climates have monthly mean temperatures above 18~$^{\circ}$C (64~$^{\circ}$F) in every month of the year and feature wet and dry seasons, as tropical savanna climates do. [1] Unlike tropical savanna climates however, a tropical monsoon climate's driest month sees less than 60 mm (2.36 in) of precipitation. [3] Also a tropical monsoon climate tends to see less variance in temperatures during the course of the year than a tropical savanna climate. [3] This climate has a driest month which nearly always occurs at or soon after the "winter" solstice for that side of the equator. [3]\\
\bottomrule
\multicolumn{2}{c}{Reference Summary} \\ \toprule
& Tropical savanna climate or tropical wet and dry climate is a type of climate that corresponds to the Köppen climate classification categories "Aw" and "As". [9] Tropical savanna climates have monthly mean temperatures above 18~$^{\circ}$C (64~$^{\circ}$F) in every month of the year and typically a pronounced dry season, with the driest month having precipitation less than 60 mm and also less than 100 -- (total annual precipitation (mm) / 25) of precipitation. [9] This latter fact is in direct contrast to a tropical monsoon climate, whose driest month sees less than 60 mm of precipitation but has more than 100 -- (total annual precipitation (mm) / 25) of precipitation. In essence, a tropical savanna climate tends to either see less rainfall than a tropical monsoon climate or have more pronounced dry seasons. [0] \\ \bottomrule
\end{tabular}
\caption{\label{tab:output:example:1} Example responses to  an \aquamuse\ query (top) for a sequence-to-sequence model which does not include  plans, two blueprint models, (abstractive and extractive), and gold standard summary.  The second block in the table shows the 5~best retrieved passages in abridged form (only the first sentence is given). Passage questions are also omitted for the sake of brevity.}
\end{table*}

\begin{table*}
  \center
  \small
\begin{tabular}{lp{13cm}} \toprule
\multicolumn{2}{c}{Abstractive Blueprint}\\ \toprule
Q: & what is a tropical monsoon climate occasionally also known as? \\
Q: &  what are the monthly mean temperatures of a tropical monsoon climate? \\
Q: & what is the average precipitation in a tropical monsoon climate's driest month?\\
Q:  & a tropical monsoon climate tends to see less variance in temperatures during the course of the year than what other type of climate? \\
Q: & what month nearly always occurs at or soon after the winter solstice for that side of the equator?\\
Q: & how many versions of a tropical monsoon climate are there?\\
Q: & what do regions with this variation of the monsoon climate typically see during the wet season? \\ \toprule
\multicolumn{2}{c}{Extractive Blueprint} \\ \toprule
Q: & what type of climate has monthly mean temperatures above 18~$^{\circ}$C (64~$^{\circ}$F) in every month of the year? \\ 
Q: & what type of climate sees less than 60 mm of precipitation in its driest month? \\
Q: & a monsoon climate tends to see more of what than a tropical savanna climate? \\ 
Q: & what type of climate has less pronounced dry seasons than tropical monsoon climates?\\
Q: & what do regions with less pronounced dry seasons typically see during the wet season? \\\toprule
\multicolumn{2}{c}{Reference Blueprint}\\ \toprule
Q:&  what is the driest month in a tropical wet and dry or savanna climate? \\
Q: &what type of climate has monthly mean temperatures above 18~$^{\circ}$C (64~$^{\circ}$F) in every month of the year?\\ 
Q: & a tropical monsoon climate's driest month sees less than what amount of precipitation? \\ 
Q: & what do tropical wet and dry or savanna climates have a pronounced dry season?\\ \bottomrule
\end{tabular}
\caption{\label{tab:blueprints:1} Abstractive and extractive blueprints for model summaries shown in Table~\ref{tab:output:example:1}. For comparison, we also include the blueprint of the reference summary.}
\end{table*}

\begin{table*}[t]
\small
\center
\begin{tabular}{lp{13cm}} 
\multicolumn{2}{c}{difference between regular file and pipe file in unix} \\ \toprule
$[$0$]$ & Tytower, I think I know what's the reason of your problem. In old Linux (a.k.a. Old Kerbel version) named pipes weren't created by mkfifo command $\dots$ \\
$[$1$]$ & It is not possible to change the backup destination for third-party tools. For this destination, the only objects created in the file system are named pipes. Named pipes occupy no space in the file system $\dots$\\
$[$2$]$ & The recovery process ends successfully, so I suppose that system correctly read all necessary recovery parameters from backup catalog. Still, it is confusing looking at the wrong log backup location in generated SQL recovery statement. $\dots$\\
$[$3$]$ &  SAP HANA supports certified third party backup tools via the backint interface. Backups are transferred via pipe from the SAP HANA database to the 3rd party backup agent, which runs on the SAP HANA database server and then sends the backups to the 3rd party backup server. $\dots$\\
$[$4$]$ &  This project shows you how to implement your own on-line radio receiver in Linux, using as major basis the mplayer application $\dots$\\
\toprule
\multicolumn{2}{c}{$-$Blueprint $+$Attribution} \\ \toprule
& In Unix-like operating systems, a file is a regular file, while a pipe is a named pipe. [0]\\ 
\bottomrule
\multicolumn{2}{c}{$+$Blueprint$_{\mathcal{A}}$ $+$Attribution} 
\\ \toprule
& Named pipes are a special type of file system object in Unix-like computer systems. [0] All Unix-like computer systems include at some point named pipes. [0]\\ \bottomrule
\multicolumn{2}{c}{$+$Blueprint$_{\mathcal{E}}$ $+$Attribution} \\ \toprule
& In Unix-like operating systems, a file is either a regular file or a named pipe. [0] Named pipes are special files which occupy no space in the file system and can be created and destroyed like any other file. [1]\\
\bottomrule
\multicolumn{2}{c}{Reference Summary} \\ \toprule
&  In Unix-like operating systems, a file is a regular file, while a pipe is a named pipe. [0]\\ \bottomrule
\end{tabular}
\caption{\label{tab:output:example:2} Example responses to  an \aquamuse\ query (top) for a sequence-to-sequence model which does not include plans, two blueprint models, (abstractive and extractive), and gold standard summary.  The second block in the table shows the 5~best retrieved passages in abridged form (only the first  couple of sentences are given). Passage questions are also omitted for the sake of brevity.}
\end{table*}

\begin{table*}
\small
  \center
\begin{tabular}{lp{13cm}} \toprule
\multicolumn{2}{c}{Abstractive Blueprint}\\ \toprule
Q: &  what are named pipes a special type of in unix-like computer systems? \\
Q: & what do all unix-like computer systems include at some point? \\
 \toprule
\multicolumn{2}{c}{Extractive Blueprint} \\ \toprule
Q: &  what type of files are created when a backup is made to a pipe? \\
Q: & named pipes occupy how much space in the file system? \\ \toprule
\multicolumn{2}{c}{Reference Blueprint}\\ \toprule
Q: & named pipes are special files that can exist where on the file system?\\
Q: & named pipe special files are made with the command mkfifo as in what ?\\ \bottomrule
\end{tabular}
\caption{\label{tab:blueprints:2} Abstractive and extractive blueprints for model summaries shown in Table~\ref{tab:output:example:2}. For comparison, we also include the blueprint of the reference summary.}
\end{table*}

\begin{table*}[t]
\small
\center
\begin{tabular}{lp{13cm}} 
\multicolumn{2}{c}{briefly explain html. what are the essential features of html} \\ \toprule
$[$0$]$ & HTML is Hyper Text Markup Language – the standard markup language for creating web applications and pages, and a cornerstone of internet programming. HTML denotes the building blocks of a web page, such as text, paragraphs, hyperlinks, images, and other items$\dots$ \\
$[$1$]$ &  HTML: Hyper Text Markup Language is the standard markup language used to create webpages and web applications. HTML represents the structure of a web page with semantic information, images, interactive forms and other objects$\dots$\\
$[$2$]$ & HyperText Markup Language or simply HTML is the markup language used by web developers to create web pages. It is written in the format of HTML elements containing tags bracketed in angle brackets like $<$html$>\dots$\\
$[$3$]$ &   HTML: HyperText Markup Language, commonly referred to as HTML, is the standard markup language used to create web pages. HTML is a standardized system for tagging text files to achieve font, color, graphic, and hyperlink effects on World Wide Web pages$\dots$\\
$[$4$]$ & HTML - HyperText Markup Language, commonly referred to as HTML, is the standard markup language used to create web pages. Web browsers can read HTML files and render them into visible or audible web pages. HTML describes the structure of a website semantically along with cues for$\dots$\\ \toprule
\multicolumn{2}{c}{$+$Blueprint$_{\mathcal{A}}$ $+$Attribution} 
\\ \toprule
&  HyperText Markup Language, commonly referred to as HTML, is the standard markup language used to create web pages. [4] Web browsers can read HTML files and render them into visible or audible web pages. [3] Web browsers do not display HTML tags and scripts, but use them to interpret the content of the page. [2] HTML describes the structure of a website semantically along with cues for presentation, making it a markup language, rather than a programming language. [4] HTML elements form the structure blocks of all web pages. [2] It allows images and objects to be embedded and can be used to create interactive forms. [2]\\ \bottomrule
\multicolumn{2}{c}{$+$Blueprint$_{\mathcal{E}}$ $+$Attribution} \\ \toprule
& HTML is HyperText Markup Language -- a standard markup computer language for representing the contents of a page of hypertext. [0] Web browsers receive HTML documents from a server and render them into visible or audible web pages. [4] HTML describes the structure of a web page semantically and originally included cues for the appearance of the document. [8]\\
\bottomrule
\multicolumn{2}{c}{Reference Summary} \\ \toprule
&  Hypertext Markup Language (HTML) is the standard markup language for creating web pages and web applications. [8] With Cascading Style Sheets (CSS) and JavaScript it forms a triad of cornerstone technologies for the World Wide Web. [5] Web browsers receive HTML documents from a web server or from local storage and render them into multimedia web pages. [2] HTML describes the structure of a web page semantically and originally included cues for the appearance of the document. [8]\\ \bottomrule
\end{tabular}
\caption{\label{tab:output:example:3} Example responses to  an \aquamuse\ query (top) for a sequence-to-sequence model which does not include  plans, two blueprint models, (abstractive and extractive), and gold standard summary.  The second block in the table shows the 5~best retrieved passages in abridged form (only the first  couple of sentences are given). Passage questions are also omitted for the sake of brevity.}
\end{table*}

\begin{table*}
  \center
  \small
\begin{tabular}{lp{13cm}} \toprule
\multicolumn{2}{c}{Abstractive Blueprint}\\ \toprule
Q:  &what is the standard markup language used to create web pages? \\
Q: &what can web browsers read and render into visible or audible web pages? \\
Q: & what do web browsers use to interpret the content of a page? \\
Q: & what describes the structure of a website semantically along with cues for presentation? \\
Q: & what do html elements form the structure blocks of?\\ 
Q: & what can be embedded in html and used to create interactive forms?  \\
 \toprule
\multicolumn{2}{c}{Extractive Blueprint} \\ \toprule
Q: & what is the standard markup language for creating web applications and pages? \\
Q: & what is a standardized system for tagging text files to achieve font, color, graphic, and hyperlink effects on world wide web pages? \\
Q: & web browsers receive html documents from a server and render them into what? \\
Q: & html describes the structure of a website semantically along with cues for what?\\ \toprule
\multicolumn{2}{c}{Reference Blueprint}\\ \toprule
Q: & what is the standard markup language for creating web pages and web applications? \\
Q: & along with html and css, what is a cornerstone technology for the world wide web? \\
Q: & web browsers receive html documents from where? \\
Q: & html originally included cues for the appearance of what?  \\ \bottomrule
\end{tabular}
\caption{\label{tab:blueprints:3} Abstractive and extractive blueprints for model summaries shown in Table~\ref{tab:output:example:3}. For comparison, we also include the blueprint of the reference summary.}
\end{table*}


\end{document}